\documentclass[11pt]{article}

\usepackage{acl}

\usepackage{times}
\usepackage{latexsym}

\usepackage[T1]{fontenc}

\usepackage[utf8]{inputenc}

\usepackage{microtype}

\usepackage{inconsolata}

\usepackage{graphicx}
\usepackage{amsmath}
\usepackage{booktabs}
\usepackage{multirow}
\usepackage[most]{tcolorbox}
\usepackage{alltt}
\title{ChartSync: A Benchmark for Visuo-Logical Cascading Chart Editing}

\author{
  \textbf{Jiakang Yu\textsuperscript{1,2,*,\textdagger}},
  \textbf{Yixuan Chai\textsuperscript{2,*}},
  \textbf{Tianci Wang\textsuperscript{3}},
  \textbf{Rihui Jin\textsuperscript{4}},
  \textbf{Guangkai Xu\textsuperscript{5}},
\\
  \textbf{Hongtao Deng\textsuperscript{1,\textdaggerdbl}},
  \textbf{Xun Zhu\textsuperscript{1}},
  \textbf{Wang Gao\textsuperscript{1,\textdaggerdbl}},
  \textbf{Xinrun Guo\textsuperscript{2}},
  \textbf{Haipang Wu\textsuperscript{2}}
\\
\\
  \textsuperscript{1}Jianghan University \quad
  \textsuperscript{2}HiThink Research \\
  \textsuperscript{3}University of Science and Technology of China \\
  \textsuperscript{4}Southeast University \quad
  \textsuperscript{5}Zhejiang University
\\
  \texttt{hongtaodeng@jhun.edu.cn, gaow@jhun.edu.cn}
}

\begin{document}
\maketitle

\begingroup
\def\thefootnote{}
\footnotetext{\textsuperscript{*}Equal contribution.}
\footnotetext{\textsuperscript{\textdagger}Work done during an internship at HiThink Research.}
\footnotetext{\textsuperscript{\textdaggerdbl}Corresponding authors.}
\endgroup

\begin{abstract}
Generative image editing models struggle with structured statistical charts when data modifications require geometric synchronization. We formalize this task as Visuo-Logical Cascading Editing (VLCE). However, existing methods remain confined to localized text substitutions and struggle with dependency-aware cascading updates. To systematically evaluate this capability, we introduce ChartSync, an expert-validated benchmark constructed via a programmatic rendering pipeline that guarantees deterministic visuo-logical coupling for the ground truth. ChartSync comprises 870 triplets across 9 chart categories and 4 task types, including 235 geometry-coupled VLCE instances that specifically test cascading text-to-geometry synchronization. We further evaluate these instances via a two-tier framework combining objective visual metrics with a vision-language model judge paradigm to assess low-level fidelity alongside multimodal comprehension and reasoning. Evaluating 14 image editing models and one code-mediated pipeline reveals a nuanced capability gap: most open-source models suffer severe drops in geometric synchronization, while only two frontier proprietary models show emerging VLCE capability, with their residual errors mainly involving semantic isolation and background corruption. Our detailed error analysis deconstructs these failure paradigms to identify core meta-abilities for guiding future multimodal architectures. The ChartSync dataset and code are publicly released at \url{https://github.com/kaka-yjk/ChartSyncCodebase}.
\end{abstract}

\section{Introduction}

\begin{figure}[t]
    \centering
    \includegraphics[width=0.9\linewidth]{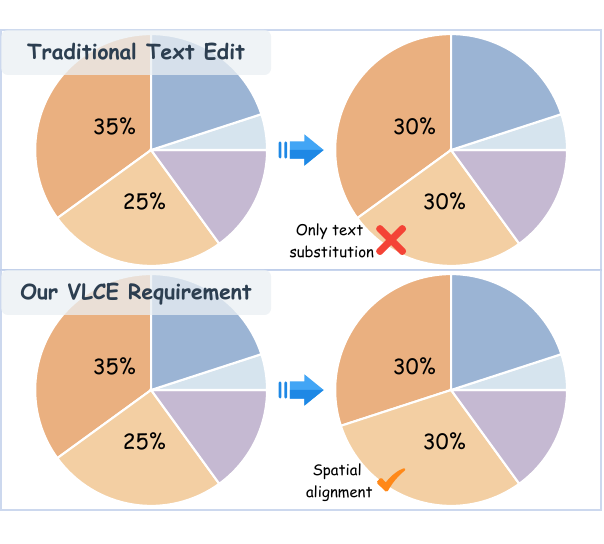}
    \caption{An illustration of the VLCE task. While traditional instruction-based editing merely substitutes text labels, our VLCE requirement demands a synchronous spatial alignment.}
    \label{fig:teaser}
\end{figure}

While instruction-based image editing has advanced rapidly in natural scenes~\cite{brooks2023instructpix2pix,zhang2023magicbrush,pan2025ice,pathiraja2025refedit,ma2024i2ebench}, extending this to statistical charts remains challenging due to precise text-geometry synchronization~\cite{gui2025texteditbench}.

Recent chart editing works often adopt a code-based paradigm, using intermediate scripts to avoid direct pixel generation~\cite{zhao-etal-2025-chartedit, yang2025chartm3}. However, this paradigm assumes source-code availability for flattened chart images and can suffer from information loss during reverse engineering~\cite{yang2025chartmimic, tang2025charts}. This motivates direct pixel-space chart editing for real-world flattened chart images.

\begin{table*}[t]
\centering
\small
\resizebox{\textwidth}{!}{
\begin{tabular}{lcccccc}
\toprule
\textbf{Benchmark} & \textbf{Domain} & \textbf{Input at Inference} & \textbf{Output / Target} & \textbf{Core Evaluated Ability} & \textbf{Value-to-Geometry Sync.} & \textbf{Sync. Metric} \\
\midrule
OmniText~\cite{gunawan2025omnitext} & Visual text & Image + mask + text/style & Image & Text content/style manipulation & No & No \\
TextEditBench~\cite{gui2025texteditbench} & Visual text & Image + instruction & Image & Reasoning-aware text editing & No & No \\
ChartEdit~\cite{zhao-etal-2025-chartedit} & Chart & Image or image + code + instruction & Code / rendered image & Code-level chart editing & Partial & No \\
ChartM3~\cite{yang2025chartm3} & Chart & Image + code + instruction & Code / rendered image & Target localization and code editing & Partial & No \\
ChartE³~\cite{li2026charte3} & Chart & Image + instruction & Image & End-to-end local/global chart editing & Partial & No \\
ChartEditVista~\cite{chen2026charteditor} & Chart & Image + instruction & Code / rendered image & Fine-grained layout/text fidelity & Partial & No \\
\textbf{ChartSync} & Chart & Image + instruction & Image & \textbf{Text-to-geometry cascading} & \textbf{Yes} & \textbf{Yes} \\
\bottomrule
\end{tabular}
}
\caption{Comparison between ChartSync and related text/chart editing benchmarks.}
\label{tab:related_comparison}
\end{table*}

Nevertheless, direct chart manipulation introduces challenges spanning multimodal comprehension, reasoning, and generation. Even simple text modifications demand precise layout alignment to prevent visual corruption~\cite{gui2025texteditbench}. Furthermore, unlike natural scenes that allow geometric flexibility, chart geometries are governed by underlying data. As argued in~\cite{li2025charts}, chart editing is a structured transformation problem rather than generic image manipulation. We formalize this task as Visuo-Logical Cascading Editing (VLCE), which requires synchronizing text modifications with geometric adjustments. As shown in Figure~\ref{fig:teaser}, modifying a pie chart's percentages requires adjusting the corresponding slices' central angles. Unfortunately, most current generative editing models are confined to localized literal substitutions, failing to grasp this underlying visuo-logical coupling. 

To fill this gap, we introduce ChartSync, an expert-validated benchmark for evaluating VLCE in statistical charts. ChartSync uses a programmatic rendering pipeline to ensure deterministic text-geometry coupling and integrates a two-tier evaluation framework covering textual fidelity, structural integrity, visuo-logical synchronization, and background preservation. Benchmarking 14 editing models plus a code-mediated pipeline reveals that VLCE capability remains highly uneven: only two frontier proprietary models show strong text-to-geometry synchronization, while most open-source and several proprietary models still exhibit substantial cascading failures. Based on these failure patterns, we identify a hierarchy of meta-abilities to guide future multimodal architectures.

In summary, our main contributions are as follows:
\begin{itemize}
\item We formally define the task of VLCE and introduce ChartSync, a focused diagnostic benchmark for direct chart editing, where the full dataset evaluates foundational chart perception and text manipulation, and the VLCE subset specifically probes dependency-aware cascading reasoning.
\item We develop a programmatic rendering pipeline to construct expert-validated editing triplets, guaranteeing deterministic visuo-logical coupling and geometric precision for the ground-truth (GT) data.
\item We integrate an automated, two-tier evaluation framework combining objective visual metrics for low-level fidelity with a vision-language model (VLM) judge paradigm to assess complex multimodal comprehension and reasoning.
\item We evaluate 14 editing models and one code-mediated baseline, revealing distinct failure paradigms and identifying core meta-abilities for robust text-to-geometry chart editing.
\end{itemize}

\section{Related Works}

\subsection{Instruction-based Image Editing}

Instruction-based image editing, pioneered by InstructPix2Pix~\cite{brooks2023instructpix2pix}, allows users to modify images via natural language commands. This approach avoids explicit masks or detailed target descriptions~\cite{hertz2022prompt, avrahami2022blended}, and has been further extended with high-quality annotated datasets~\cite{zhang2023magicbrush,ye2026imgedit} and multimodal visual guidance~\cite{fu2024guiding}.

However, current instruction-based editing models focus on natural scenes and struggle with the rigid visual structures and data dependencies of statistical charts~\cite{zhao-etal-2025-chartedit, yang2025chartm3}. Chart editing requires strict dependency-aware reasoning for geometric synchronization, a challenge emphasized by recent studies \cite{yang2025chartmimic, tang2025charts}. Manipulating such structures demands models equipped with dependency-aware visuo-logical reasoning to achieve geometric synchronization, alongside strict semantic isolation to preserve background integrity. These represent critical needs echoed by recent efforts~\cite{he2025diffthinker, dai2026endocot}.

\subsection{Text and Chart Editing Benchmarks}

\begin{figure*}[t]
    \centering
    \includegraphics[width=\textwidth]{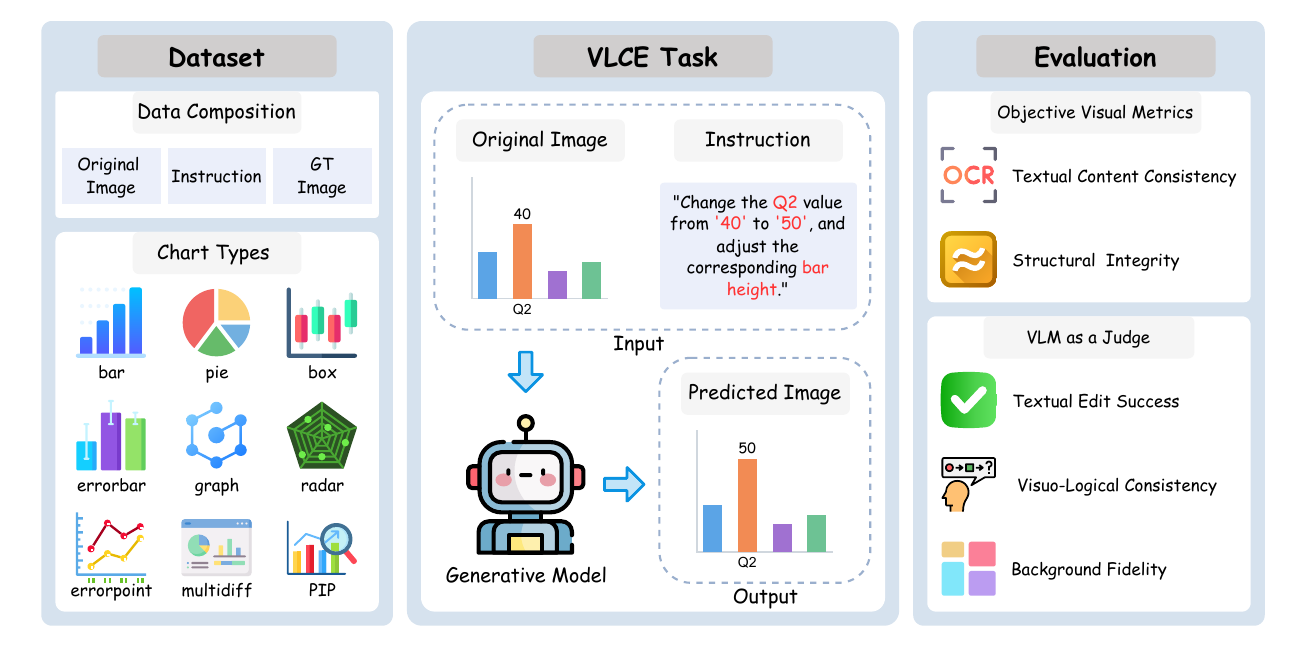}
    \caption{An overview of ChartSync, consisting of a dataset, a VLCE task formulation, and an evaluation framework.}
    \label{fig:overview}
\end{figure*}

Recent text-centric benchmarks evaluate fine-grained textual manipulation in images. OmniText~\cite{gunawan2025omnitext} studies controllable text-image manipulation, including text removal, insertion, editing, rescaling, repositioning, and style control. TextEditBench~\cite{gui2025texteditbench} further evaluates reasoning-aware text editing across diverse visual documents with criteria such as text accuracy, visual consistency, layout preservation, and semantic expectation. These works highlight the difficulty of editing embedded text under style and layout constraints, but do not target chart-specific data-to-geometry dependencies.

Chart editing has also received increasing attention. ChartEdit~\cite{zhao-etal-2025-chartedit} evaluates MLLMs on chart editing through code generation, with both chart-level and code-level evaluation. ChartM3~\cite{yang2025chartm3} introduces multimodal chart editing with textual descriptions and visual indicators, requiring models to associate highlighted chart regions with code-level elements. More recently, ChartE$^3$~\cite{li2026charte3} proposes an end-to-end chart editing benchmark covering both local appearance edits and global data-centric transformations. ChartEditVista~\cite{chen2026charteditor} further scales chart editing to thousands of image-instruction samples without requiring original chart code as input, and proposes fine-grained layout and text metrics.

Despite these advances, existing benchmarks primarily evaluate text manipulation, target localization, code-level edit correctness, layout fidelity, textual fidelity, or overall chart editing faithfulness. They do not explicitly isolate value-to-geometry synchronization as the core evaluation target, where a textual value update must trigger a coupled geometric deformation in the rendered chart. ChartSync is therefore positioned as a focused diagnostic benchmark for pixel-space VLCE: it uses image-only editing inputs, constructs deterministic GT through programmatic rendering, and introduces VLCS to directly evaluate whether the edited geometry is synchronized with the requested textual value change. Table~\ref{tab:related_comparison} summarizes this distinction.

\section{ChartSync}

As depicted in Figure~\ref{fig:overview}, ChartSync consists of three components: a curated dataset, a formal task formulation of VLCE, and an evaluation framework.

\subsection{Task Definition}

Formally, we define a structured chart as a visuo-logical graph $C=(V,G,S)$, where $V=\{v_i\}$ denotes textual or numerical values, $G=\{g_j\}$ denotes geometric primitives, and $S$ captures the dependency between value changes and geometric deformations. The visuo-logical constraint is defined as:

\begin{equation}
    \Delta G = S(\Delta V).
\end{equation}

Given an original chart image $I_{ori}$ and an instruction $T_{inst}$ specifying $\Delta V$, an editing model generates:
\begin{equation}
    I_{pred} = \mathcal{M}(I_{ori}, T_{inst}).
\end{equation}

\begin{figure*}[t]
    \centering
    \includegraphics[width=0.9\textwidth]{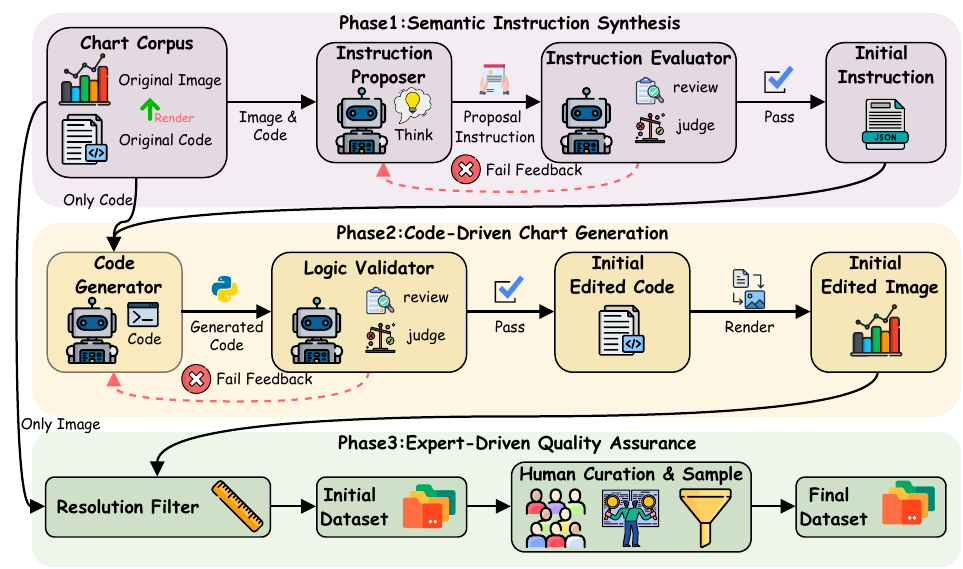} 
    \caption{The three-stage data construction pipeline of ChartSync. Phase 1 generates and evaluates initial editing instructions, Phase 2 produces GT images through code modification and logic validation, and Phase 3 performs expert quality assurance.}
    \label{fig:pipeline}
\end{figure*}

A valid VLCE output satisfies three conditions: 1) textual realization, where the target values reflect $\Delta V$; 2) geometric synchronization, where geometry follows $\Delta G=S(\Delta V)$; and 3) non-target preservation, where unedited elements remain unchanged. Thus, unlike standard instruction-based editing, VLCE requires models to infer and apply data-to-graphic dependencies rather than merely modifying visible text.

\subsection{Data Construction}

As illustrated in Figure~\ref{fig:pipeline}, we design a three-stage pipeline integrating automated generation with human curation, utilizing original rendering codes from ChartMimic~\cite{yang2025chartmimic}.

\paragraph{Phase 1: Semantic Instruction Synthesis.} A VLM-based \textbf{Instruction Proposer} first extracts semantic key-value pairs to generate initial instructions. Subsequently, an \textbf{Instruction Evaluator} filters proposals based on format, localizability, and training value. The evaluator intentionally retains localized text-only edits that implicitly involve chart dependencies, such as altering a numerical value without specifying its corresponding geometric scaling, thereby requiring downstream agents to infer the implicit cascading transformation.

\paragraph{Phase 2: Code-Driven Chart Generation.} A \textbf{Code Generator} powered by a Large Language Model (LLM) modifies the Python scripts with minimal edits while triggering causally related structural updates. A \textbf{Logic Validator} forms a feedback loop with the generator to verify syntax validity and instruction alignment. Inconsistent cases are iteratively corrected before rendering the final ground-truth images $I_{gt}$.

\paragraph{Phase 3: Expert-Driven Quality Assurance.} An automated Resolution Filter removes samples whose rendered resolutions are inconsistent with the original images. Three artificial intelligence domain experts independently inspect each remaining triplet and its corresponding code under a unanimous acceptance protocol. Detected errors, artifacts, or ambiguities are corrected through instruction refinement and chart re-rendering. After quality control and stratified sampling, we retain 870 expert-validated instances with unambiguous instructions and unique target images. Detailed expert qualifications, QA protocols, correction statistics, and retention statistics are provided in Appendix~\ref{sec:appendix_expert_qa}.

\subsection{Dataset Statistics}

To cover diverse data analysis scenarios, the dataset spans 9 chart categories and 4 progressively challenging task types, as depicted in Figure~\ref{fig:distribution}. Detailed dataset statistics are provided in Appendix~\ref{sec:appendix_data_stat}.

Regarding visual domains, ChartSync covers both standard statistical charts and advanced composite layouts. Standard formats include bar charts, box plots, errorpoint charts, graphs, radar charts, errorbar charts, and pie charts, while advanced layouts include Plot-in-Plot (PIP) and multidiff charts to evaluate models under dense and structured visual conditions.

Tasks are categorized into four types: \textit{single text edit}, \textit{multiple text edit}, \textit{single VLCE}, and \textit{multiple VLCE}, explicitly decoupling element quantity from structural updates. Text-only tasks evaluate foundational chart comprehension and text manipulation, while VLCE tasks require models to infer value-geometry dependencies and perform synchronized updates. Among the 870 triplets, 235 are geometry-coupled VLCE instances, including 179 single VLCE and 56 multiple VLCE samples. Therefore, the full dataset evaluates direct chart editing broadly, whereas cascading reasoning claims are based specifically on the VLCE subset.

\begin{figure}[t]
    \centering
    \includegraphics[width=\linewidth]{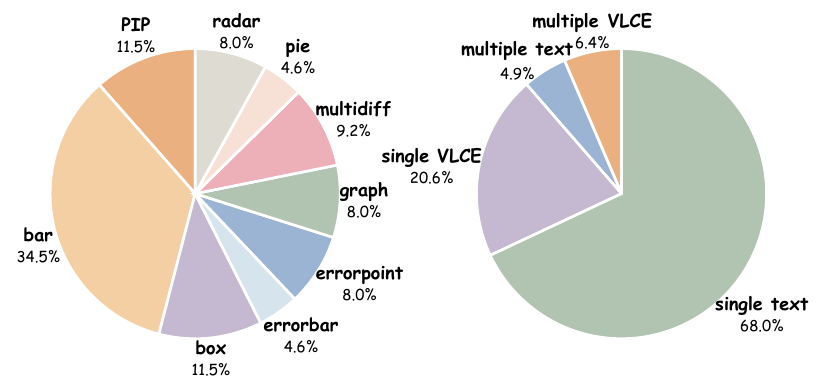}
    \caption{Data distribution of chart categories and task types in ChartSync.}
    \label{fig:distribution}
\end{figure}

\subsection{Evaluation Metrics}

We evaluate VLCE via a two-tier framework combining objective visual metrics with a VLM-as-a-Judge paradigm, as illustrated in Figure~\ref{fig:overview}.

\subsubsection{Objective Visual Metrics}

This component measures textual and structural fidelity using character-level and pixel-level metrics.

\paragraph{OCR $\text{F}_1$ for Textual Content Consistency.}
We evaluate full-image textual fidelity using a character-level $\text{F}_1$ score computed from text sequences extracted via Optical Character Recognition (OCR). Specifically, we normalize predicted and GT texts by removing whitespace and lowercasing, then compute character-level precision and recall based on the longest common subsequence between them. The final $\text{F}_1$ score is computed as the harmonic mean of character-level precision and recall. This metric captures both correct text editing and preservation of unchanged textual content.

\paragraph{SSIM for Structural Integrity.}
The Structural Similarity Index (SSIM)~\cite{wang2004image} measures global visual consistency between $I_{pred}$ and $I_{gt}$ across luminance, contrast, and structure, capturing structural changes and potential artifacts in non-edited regions.

\subsubsection{VLM as a Judge}

While objective metrics capture low-level fidelity, they are insufficient for evaluating complex multimodal comprehension and reasoning. Therefore, we introduce a VLM-as-a-Judge paradigm, where a proprietary VLM performs holistic evaluation given the triplet \textit{[Original, GT, Prediction]}. 

\paragraph{TESR for Textual Edit Success.}
Textual Edit Success Rate (TESR) is a discrete score in $\{0.0, 0.2, 0.4, 0.6, 0.8, 1.0\}$ that measures the execution of the instructed text modification, evaluating both OCR-level semantic correctness and spatial precision.

\paragraph{VLCS for Visuo-Logical Consistency.}
For single VLCE and multiple VLCE tasks, Visuo-Logical Consistency Score (VLCS) evaluates whether the edited visual elements correctly reflect the underlying data modifications. Scored discretely in $\{0.0, 0.25, 0.5, 1.0\}$, this metric focuses exclusively on the geometric regions directly coupled to the edited text, penalizing models that fail to synchronize data values with visual representations. 

\paragraph{BFS for Background Fidelity.}
Background Fidelity Score (BFS) measures the preservation of unmodified chart components. Scored discretely in $\{0.0, 0.25, 0.5, 0.75, 1.0\}$, it explicitly assesses untargeted areas for visual artifacts, color shifts, or structural collapse, capturing global layout coherence.

To aggregate these insights, \textbf{VLM Avg} is computed as the macro-average of the three VLM-derived metric means. TESR and BFS are computed over all applicable samples, while VLCS is evaluated only on the 235 VLCE instances without assigning scores to text-only samples. The \textbf{Overall Score} is the equal average of five metric-level scores: OCR $\text{F}_1$, SSIM, TESR, VLCS, and BFS, and should be interpreted as a metric-level summary rather than a sample-level average.

\section{Experiments}

\begin{table*}[t]
\centering
\resizebox{\textwidth}{!}{
\begin{tabular}{l | cc | ccc | cc}
\toprule
\multirow{2}{*}{\textbf{Model}} & \multicolumn{2}{c|}{\textbf{Objective Visual Metrics}} & \multicolumn{3}{c|}{\textbf{VLM as a Judge}} & \multirow{2}{*}{\textbf{VLM Avg}} & \multirow{2}{*}{\textbf{Overall}} \\
\cmidrule{2-3} \cmidrule{4-6}
 & \textbf{OCR $\text{F}_1$} & \textbf{SSIM} & \textbf{TESR} & \textbf{VLCS} & \textbf{BFS} & & \\
\midrule
Code-Mediated Pipeline & \textbf{88.76} & 54.83 & 93.38 & 28.51 & 37.44 & 53.11 & 60.58 \\
\midrule
\multicolumn{8}{c}{\textit{Proprietary Models}} \\
Qwen-Image-2.0-Pro~\cite{zhao2026qwenimage20technicalreport}             & 78.24 & \textbf{91.30} & 74.05 & 24.26 & 59.83 & 52.71 & 65.54 \\
SeeDream-5.0-Lite~\cite{seedream2025seedream}              & 85.30 & 77.29 & 85.61 & 37.66 & 69.60 & 64.29 & 71.09 \\
Wan2.7-Image-Pro~\cite{mao2026wan}               & 81.16 & 79.68 & 82.62 & 56.38 & 72.50 & 70.50 & 74.47 \\
GPT-Image-2\footnotemark[1]
& 83.88 & 81.34 & 88.90 & 74.47 & 80.29 & 81.22 & 81.78 \\
Nano Banana Pro\footnotemark[2]
& 86.02 & 83.20 & \textbf{96.25} & \textbf{83.71} & \textbf{89.60} & \textbf{89.85} & \textbf{87.76} \\
\midrule
\multicolumn{8}{c}{\textit{Open-Source Models}} \\
DeepGen-1.0~\cite{wang2026deepgen}                    & 43.51 & 26.21 & 13.72 & 8.72  & 25.89 & 16.11 & 23.61 \\
Intern-VL-U~\cite{tian2026internvl}                    & 49.66 & 24.17 & 25.91 & 12.02 & 24.14 & 20.69 & 27.18 \\
Step1X-Edit-v1p2~\cite{liu2025step1x-edit}               & 64.79 & 77.37 & 27.24 & 10.43 & 49.68 & 29.12 & 45.90 \\
LongCat-Image-Edit~\cite{team2025longcat}             & 62.90 & 74.66 & 36.39 & 13.19 & 49.11 & 32.90 & 47.25 \\
FLUX.2-Klein-Base-9B~\cite{flux-2-2025}          & 72.18 & 69.28 & 43.47 & 12.55 & 55.60 & 37.21 & 50.62 \\
FireRed-Image-Edit-1.0~\cite{team2026firered}         & 71.39 & 76.28 & 48.09 & 12.02 & 50.29 & 36.80 & 51.61 \\
FireRed-Image-Edit-1.1~\cite{team2026firered}         & 74.09 & 77.58 & 60.23 & 12.77 & 56.41 & 43.14 & 56.22 \\
Qwen-Image-Edit~\cite{wu2025qwen}                & 65.63 & 78.36 & 43.54 & 11.49 & 54.71 & 36.58 & 50.75 \\
Qwen-Image-Edit-2511~\cite{wu2025qwen}           & 76.93 & 90.02 & 61.81 & 13.83 & 58.88 & 44.84 & 60.29 \\
\bottomrule
\end{tabular}
}
\caption{Results of 14 image editing models and one code-mediated pipeline on the ChartSync benchmark.}
\label{tab:main_results}
\end{table*}

\subsection{Experimental Setup}

All experiments are conducted on NVIDIA A100 GPUs. We evaluate 14 instruction-based image editing models to assess their VLCE capabilities, covering both proprietary and open-source image editing models. In addition, we include a code-mediated baseline to examine whether a chart-to-code paradigm can solve ChartSync when the original source code is unavailable. This pipeline first reconstructs executable plotting code from the input chart image using a VLM, then edits the reconstructed code according to the instruction, and finally re-renders the edited chart. The resulting image is evaluated under the same ChartSync protocol as direct image editing models. We apply a direct prompting strategy, providing the models with the original chart image and direct instructions. Comprehensive hyperparameter configurations for all evaluated baselines, as well as the LLMs and VLMs utilized in our automated pipeline, are detailed in Appendix~\ref{sec:appendix_configs}. Additionally, the exact system prompts guiding the data construction and the VLM-as-a-Judge evaluation are provided in Appendix~\ref{sec:appendix_prompts}.

\footnotetext[1]{
\url{https://developers.openai.com/api/docs/models/gpt-image-2}.
Model identifier: \texttt{gpt-image-2-2026-04-21}. Accessed July 2026.
}

\footnotetext[2]{
\url{https://ai.google.dev/gemini-api/docs/models/gemini-3-pro-image}.
Model identifier: \texttt{gemini-3-pro-image}. Accessed July 2026.
}

\subsection{Main Results}

Table~\ref{tab:main_results} presents the quantitative performance of 14 image editing models and one code-mediated pipeline on ChartSync, yielding several key findings.

\paragraph{Severe Failure in Cascading Updates.}
A sharp performance degradation occurs when models transition from localized text modifications to geometric synchronization. Across the benchmark, most models exhibit a substantial gap between TESR and VLCS. For example, Qwen-Image-Edit-2511 achieves a TESR of 61.81 but plummets to a VLCS of 13.83. This discrepancy demonstrates that current generative editing methods remain largely confined to localized literal substitutions, failing to comprehend the underlying data logic or propagate textual modifications into geometric synchronization.

\paragraph{Code-Mediated Editing Is Not a Free Solution.}
The code-mediated baseline achieves high OCR $\text{F}_1$ and TESR scores of 88.76 and 93.38, indicating that chart-to-code reconstruction followed by code editing can often perform textual modifications. However, its SSIM, VLCS, and BFS drop to 54.83, 28.51, and 37.44, respectively, revealing substantial information loss during image-to-code reconstruction. These results show that code-mediated pipelines remain limited when source code is unavailable, motivating direct pixel-space chart editing for real-world scenarios.

\begin{figure*}[t]
    \centering
    \includegraphics[width=\textwidth]{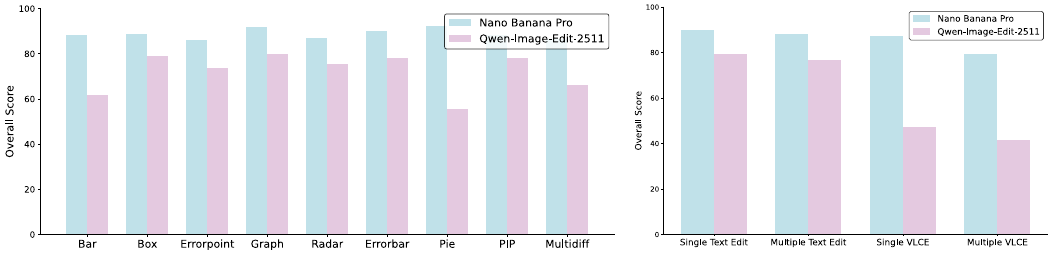}
    \caption{Performance comparison between Nano Banana Pro and Qwen-Image-Edit-2511. The left panel illustrates the overall scores across 9 chart categories, while the right panel details the performance breakdown by task complexity.}
    \label{fig:compare}
\end{figure*}

\paragraph{Frontier Models Show Emerging VLCE Capability.}
The VLCS results reveal a more nuanced picture than a simple proprietary-versus-open-source dichotomy. Nano Banana Pro and GPT-Image-2 achieve strong VLCS scores of 83.71 and 74.47, respectively, indicating emerging capability in text-to-geometry synchronization. However, the remaining proprietary models still range from 24.26 to 56.38, and the best open-source model, Qwen-Image-Edit-2511, reaches only 13.83. This suggests that reliable VLCE is not a general property of proprietary models, but is currently concentrated in a few frontier systems.

\paragraph{The Illusion of Structural Fidelity.}
While high SSIM suggests global layout preservation, BFS reveals an illusion of structural fidelity. Despite SSIM values exceeding 90 for Qwen-Image-2.0-Pro and Qwen-Image-Edit-2511, most models yield moderate BFS scores, with open-source models averaging around 50. This discrepancy arises from a deficit in fine-grained semantic comprehension. Without precise structural grounding, these baselines struggle to accurately isolate the target text and its coupled geometries. Consequently, their edits frequently bleed into untargeted regions, causing collateral damage to essential structures such as axes, gridlines, or adjacent data marks.

\section{Analysis and Discussion}

\subsection{Performance Across Chart Categories and Task Complexity}

To understand how visual structures influence editing capabilities, we compare Nano Banana Pro and Qwen-Image-Edit-2511 across chart categories, as illustrated in Figure~\ref{fig:compare}. While the proprietary model maintains robust performance across categories, the open-source model exhibits larger fluctuations, particularly on bar, pie, and multidiff charts. These results highlight the difficulty of dependency-aware geometric synchronization and semantic isolation in complex chart structures.

Regarding task complexity, increasing editing elements from single to multiple causes only marginal degradation for both models. This suggests that multi-target editing is not the primary bottleneck; instead, geometric synchronization remains the dominant challenge.

\subsection{Error Analysis and Meta-Abilities}

\begin{figure}[t]
    \centering
    \includegraphics[width=\columnwidth]{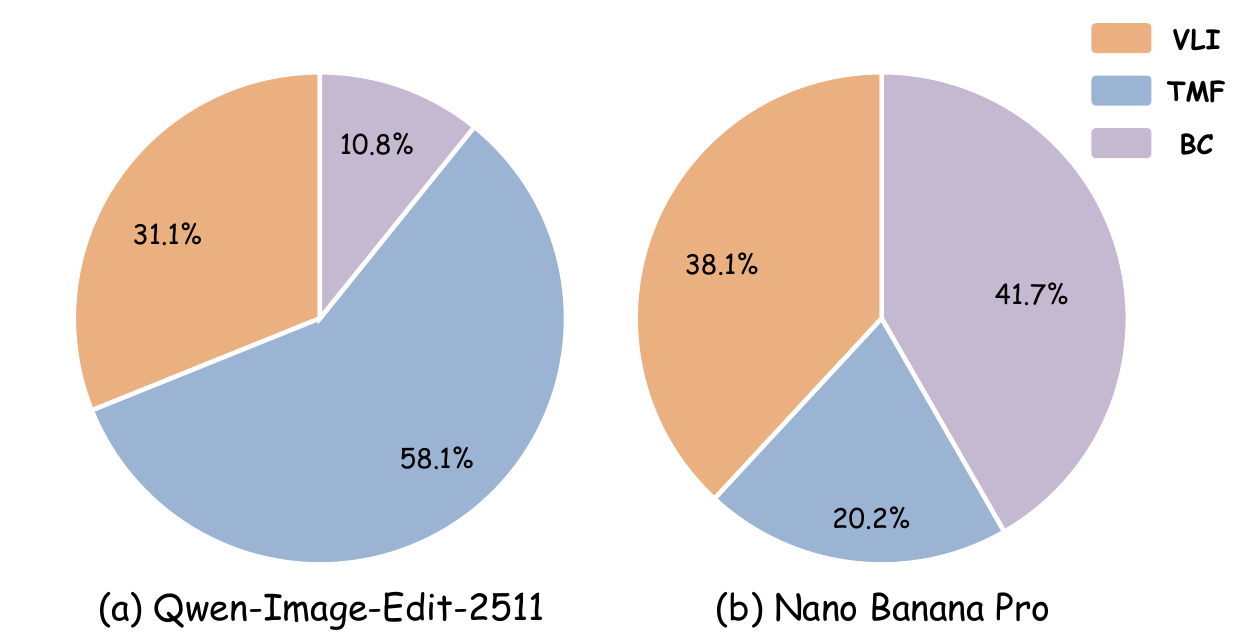}
    \caption{Distribution of primary error occurrences within the VLCE subset.}
    \label{fig:error_pie}
\end{figure}

To isolate critical bottlenecks in generative chart editing, we conduct a representative diagnostic comparison between the top-performing open-source baseline, Qwen-Image-Edit-2511, and the leading proprietary model, Nano Banana Pro, within the VLCE subset. This analysis characterizes representative failure modes rather than providing an exhaustive taxonomy over all models. As illustrated in Figure~\ref{fig:error_pie}, we treat the evaluation of degraded images as a multi-label classification task and categorize all observed error occurrences into three distinct types. Textual Manipulation Failure (TMF) occurs when a model fails to accurately locate or render the target text. Visuo-Logical Inconsistency (VLI) arises when numerical text changes fail to propagate to the associated quantitative geometry. Background Corruption (BC) represents cases where untargeted background elements are corrupted despite correct target editing. For Qwen-Image-Edit-2511, TMF at 58.1 and VLI at 31.1 dominate the errors, indicating failures in both text rendering and dependency-aware synchronization. In contrast, Nano Banana Pro largely overcomes foundational text failures, with remaining errors concentrated on BC at 41.7 and VLI at 38.1. 

Drawing insights from these distinct failure paradigms, we deconstruct the model requirements into three hierarchical meta-abilities. First, \textbf{foundation perception} entails semantic grounding to map abstract instructions to specific visual objects, coupled with spatial layout awareness to prevent modified elements from overlapping with surrounding visual components. Second, \textbf{synchronization and reasoning} enables cascading editing through visual-quantitative alignment between numerical changes and geometric updates, together with structural reasoning over causally related elements. Third, \textbf{high-fidelity generation} requires artifact-free rendering, geometric preservation, and semantic isolation to prevent visual bleeding during localized edits.

To further illustrate these failure modes, Appendix~\ref{sec:appendix_qualitative} provides four qualitative scenarios. Specifically, case 1 demonstrates the breakdown of foundational perception through catastrophic TMF and structural disintegration. Next, case 2 illustrates the synchronization and reasoning bottleneck, highlighting how absolute VLI renders open-source graphics entirely static while proprietary baselines achieve precise text-geometry synchronization. Furthermore, case 3 illustrates a residual high-fidelity generation failure, showing that semantic isolation can still break down in specific cases even when literal text edits are correctly executed. Finally, case 4 marks the non-saturated boundary of the benchmark, revealing divergent geometric scaling errors when architectures encounter highly coupled visual-data dependencies.

\subsection{Correlation With Human Evaluation}

To validate the reliability of our two-tier evaluation framework, we conduct a blind human evaluation on 200 generated results sampled across different models. Three independent experts assess each prediction on a 0--100 scale based on criteria covering textual fidelity, geometric synchronization, and background preservation. We compute the Intraclass Correlation Coefficient (ICC) to measure inter-annotator agreement, achieving an overall ICC of 0.87, indicating high consistency among human judgments.

We average the expert scores as the final human judgments and compute the Pearson correlation coefficient $r$ and Spearman's rank correlation $\rho$ against our automated metrics. As presented in Table~\ref{tab:human_correlation}, all automated metrics demonstrate strong positive correlations with human assessments at a significance level of $p < 0.001$. Notably, the VLM-as-a-Judge metrics exhibit significantly higher alignment than traditional objective metrics. In particular, VLCS achieves the highest correlation of $r = 0.892$, indicating that VLCS effectively captures human-perceived geometric synchronization quality. These results indicate that our VLM-as-a-Judge protocol is not used as an unvalidated oracle, but is calibrated against blind human annotations.

\begin{table}[h]
    \centering
    \small
    \begin{tabular}{lcc}
        \toprule
        \textbf{Metric} & \textbf{Pearson ($r$)} & \textbf{Spearman ($\rho$)} \\
        \midrule
        OCR $\text{F}_1$ & 0.814 & 0.795 \\
        SSIM & 0.695 & 0.688 \\
        TESR & 0.867 & 0.852 \\
        VLCS & 0.892 & 0.884 \\
        BFS & 0.845 & 0.830 \\
        \midrule
        \textbf{Overall Score} & \textbf{0.915} & \textbf{0.903} \\
        \bottomrule
    \end{tabular}
    \caption{Correlation between our automated multi-dimensional evaluation metrics and human expert judgments.}
    \label{tab:human_correlation}
\end{table}

\section{Conclusion}

In this research, we introduced ChartSync, a benchmark evaluating generative models on VLCE within statistical charts. We constructed 870 expert-validated editing triplets using a rigorous programmatic rendering pipeline. Alongside this dataset, we integrated an automated, two-tier evaluation framework that decouples low-level visual fidelity from complex multimodal comprehension and reasoning. Our evaluation of 14 image editing models and one code-mediated pipeline exposes a nuanced capability landscape: most models remain severely bottlenecked by foundational text rendering and dependency-aware reasoning, while only a few frontier proprietary models show strong cascading synchronization, with semantic isolation remaining a residual failure mode in some cases. By deconstructing these distinct failure paradigms into core meta-abilities, we anticipate ChartSync will serve as a critical catalyst for advancing robust structural document understanding and guiding the development of future multimodal architectures capable of precise geometric synchronization.

\section*{Limitations}

Our work presents five primary limitations: (1) Reliance on ``VLM-as-a-Judge'': While highly scalable, this paradigm may introduce subtle biases or inaccuracies compared to strict human evaluation. Future work could incorporate multi-judge ensembles or cross-family judge models to further reduce evaluator-specific preferences and improve robustness. (2) Focus on standard statistical charts: Extending this framework to more complex document scenarios remains an open and important direction to explore. (3) Limited dataset scale: Developing a larger evaluation benchmark in the future could help uncover more nuanced findings and comprehensively evaluate multimodal editing architectures. (4) Limited scope of error taxonomy: our detailed error analysis focuses on two representative models to diagnose major failure modes, rather than exhaustively categorizing all 14 evaluated image editing models. A full model-wise taxonomy is an important direction for future analysis. (5) For proprietary models, full training-data auditing is impossible, and we therefore cannot completely rule out exposure to similar chart-related editing patterns.

\section*{Acknowledgments}
This work is supported by National Natural Science Foundation of China (NSFC, No.62276196), Industry-University-Research Project of Wuhan Education Bureau (No.CXY202208) and The Research Fund of Jianghan University (Grant No.2022XKZK10).

\bibliography{anthology,custom}

\appendix

\section{Expert Quality Assurance}
\label{sec:appendix_expert_qa}

The final candidate triplets were not accepted solely by the LLM-based logic validator. After code-level generation and automatic logic validation, 935 candidate triplets entered the expert quality assurance stage. Three Ph.D.-level AI researchers independently inspected each candidate triplet and its corresponding code. The experts checked whether the instruction was unambiguous, whether the target text was correctly modified, whether the coupled geometric update was consistent with the data change, and whether the rendered GT image contained visual artifacts or unintended layout shifts.

We adopted a unanimous-accept protocol: a triplet was retained only when all three experts agreed that the instruction, edited code, and rendered GT image were valid. In the initial QA pass, 258 candidate triplets were flagged for further correction, corresponding to a correction rate of 27.6\%. The average pairwise agreement among the experts was 95.4\%. Flagged candidates were manually corrected through instruction refinement, code adjustment, and chart re-rendering. After expert curation and stratified sampling, we retained 870 expert-validated triplets, corresponding to a final retention rate of 93.0\%. The detailed QA statistics are summarized in Table~\ref{tab:expert_qa_statistics}.

\begin{table}[t]
    \centering
    \footnotesize
    \setlength{\tabcolsep}{4pt}
    \renewcommand{\arraystretch}{1.15}
    \begin{tabular}{p{0.56\linewidth}p{0.34\linewidth}}
        \toprule
        \textbf{Item} & \textbf{Value} \\
        \midrule
        Expert-QA candidates & 935 \\
        Experts & 3 Ph.D \\
        QA protocol & Unanimous accept \\
        Initially flagged candidates & 258 \\
        Correction rate & 27.6\% \\
        Pairwise expert agreement & 95.4\% \\
        Final retained triplets & 870 \\
        Final retention rate & 93.0\% \\
        \bottomrule
    \end{tabular}
    \caption{Expert quality assurance statistics for ChartSync construction.}
    \label{tab:expert_qa_statistics}
\end{table}

\section{Detailed Dataset Statistics}
\label{sec:appendix_data_stat}

This appendix provides the precise quantitative breakdown of the ChartSync dataset. Tables~\ref{tab:dist_category} and \ref{tab:dist_task} detail the distributions of the 870 expert-curated editing triplets, independently categorized by the nine distinct chart types and the four task types. The category distribution balances common statistical charts and complex composite layouts, while the task distribution covers both foundational text manipulation and geometry-coupled VLCE reasoning. This controlled design ensures broad structural diversity and enables systematic evaluation of VLCE capabilities.

\begin{table}[h]
    \centering
    \renewcommand{\arraystretch}{1.15}
    \begin{tabular}{lc}
        \toprule
        \textbf{Chart Category} & \textbf{Count} \\
        \midrule
        Bar & 300 \\
        PIP & 100 \\
        Box & 100 \\
        Multidiff & 80 \\
        Errorpoint & 70 \\
        Graph & 70 \\
        Radar & 70 \\
        Errorbar & 40 \\
        Pie & 40 \\
        \midrule
        \textbf{Total} & \textbf{870} \\
        \bottomrule
    \end{tabular}
    \caption{Triplet distribution by chart category.}
    \label{tab:dist_category}
\end{table}

Bar charts constitute the largest subset because they are among the most prevalent statistical visualizations and provide a representative setting for evaluating value-to-geometry dependencies. The remaining chart categories cover diverse visual structures, including statistical distributions, relational graphs, and composite layouts such as PIP and multidiff charts.

\begin{table}[h]
    \centering
    \renewcommand{\arraystretch}{1.15}
    \begin{tabular}{lc}
        \toprule
        \textbf{Task Type} & \textbf{Count} \\
        \midrule
        Single Text Edit & 592 \\
        Single VLCE & 179 \\
        Multiple Text Edit & 43 \\
        Multiple VLCE & 56 \\
        \midrule
        \textbf{Total} & \textbf{870} \\
        \bottomrule
    \end{tabular}
    \caption{Triplet distribution by task complexity.}
    \label{tab:dist_task}
\end{table}

The task distribution separates textual manipulation complexity from dependency-aware geometric reasoning. Text-only tasks evaluate foundational chart comprehension and text editing ability, while VLCE tasks require models to infer value-geometry dependencies and perform synchronized updates. The controlled allocation across task types avoids excessive concentration on a single editing pattern and supports systematic evaluation of cascading reasoning capability.

\section{Model Configurations}
\label{sec:appendix_configs}

\subsection{Baseline Inference Configurations}

In this section, we detail the hyperparameter configurations for all baseline models evaluated on ChartSync. To ensure a fair and reproducible comparison, we strictly adhere to the default parameter settings officially recommended by the respective authors or technical reports.

For open-source diffusion-based models, we primarily configure two critical hyperparameters: the Classifier-Free Guidance (CFG) scale and the number of denoising inference steps, as summarized in Table~\ref{tab:model_configs}. For proprietary models accessed via API, we utilize their default black-box inference settings to reflect their standard performance in real-world applications. For the code-mediated baseline, we use GPT-5.5\footnote{\url{https://developers.openai.com/api/docs/models/gpt-5.5}} to perform image-to-code reconstruction, instruction-guided code editing, and re-rendering, and evaluate the final rendered images under the same ChartSync protocol.

\begin{table}[h]
    \centering  
    \small 
    \begin{tabular}{lcc}
        \toprule
        \textbf{Model} & \textbf{Guidance Scale} & \textbf{Steps} \\
        \midrule
        FireRed-Image-Edit-1.0 & 4.0 & 40 \\
        FireRed-Image-Edit-1.1 & 4.0 & 40 \\
        Qwen-Image-Edit        & 4.0 & 40 \\
        Qwen-Image-Edit-2511   & 4.0 & 40 \\
        Intern-VL-U            & 3.5 / 1.5$^\dagger$ & 20 \\
        LongCat-Image-Edit     & 4.5 & 50 \\
        Step1X-Edit-v1p2       & 6.0 & 50 \\
        DeepGen-1.0            & 4.0 & 50 \\
        FLUX.2-Klein-Base-9B   & 4.0 & 50 \\
        \bottomrule
    \end{tabular}
    \caption{Inference configurations for open-source models. $^\dagger$Intern-VL-U employs a Dual-Guidance CFG mechanism to decouple image reference and text instruction.}
    \label{tab:model_configs}
\end{table}

\subsection{Data Construction and Evaluation Configurations}

In addition to the baseline models, we specify the configurations for the LLMs and VLMs utilized throughout our automated data construction pipeline and the VLM-as-a-Judge evaluation framework. For these reasoning-intensive tasks, we primarily adjust the sampling temperature to balance generation diversity and logical rigor. The detailed settings are summarized in Table~\ref{tab:pipeline_configs}.

\begin{table}[h]
    \centering
    \small
    \begin{tabular}{llc}
        \toprule
        \textbf{Role} & \textbf{Model} & \textbf{Temperature} \\
        \midrule
        Instruction Proposer & GPT-5.4 & 1.0 \\
        Instruction Evaluator & GPT-5.4 & 1.0 \\
        Code Generator & GPT-5.4 & 0.2 \\
        Logic Validator & GPT-5.4 & 0.1 \\
        OCR Extractor & GPT-5.4 & 0.0 \\
        Judge & Gemini-3.1-Pro & 0.1 \\
        \bottomrule
    \end{tabular}
    \caption{Model configurations for data construction and evaluation. Higher temperatures encourage diversity in instruction synthesis, while lower temperatures ensure deterministic reasoning during logic validation and evaluation.}
    \label{tab:pipeline_configs}
\end{table}

\section{Prompts}
\label{sec:appendix_prompts}

\subsection{Data Generation Prompts}
\label{subsec:data_gen_prompts}

We present the exact system prompts used in our data construction pipeline as follows. For phase 1, the prompts for the Instruction Proposer and the Instruction Evaluator are detailed in Figure~\ref{fig:phase1_prompts_proposer} and Figure~\ref{fig:phase1_prompts_evaluator}, respectively. For phase 2, the prompts for the Code Generator and the Logic Validator are provided in Figure~\ref{fig:phase2_prompt_generator} and Figure~\ref{fig:phase2_prompt_validator}, respectively.

\begin{figure*}[t]
\centering
\begin{tcolorbox}[
    enhanced, boxrule=0.8pt, colback=white, colframe=blue!50!black, rounded corners, arc=3pt,
    width=0.98\linewidth, fontupper=\ttfamily\scriptsize, title={Phase 1: Instruction Proposer Prompt}, 
    colbacktitle=blue!50!black, coltitle=white, fonttitle=\bfseries\sffamily\scriptsize,
    left=5pt, right=5pt, top=5pt, bottom=5pt
]
\begin{alltt}
You are an expert in chart data augmentation.
Please carefully analyze the provided Python code and the chart image (generated by running the provided code). 
Based on a logical "Key-Value pair", select ONE suitable target text ('old_value') to edit, 
with the goal of generating an 'instruction'.

[Task]
1. Identify logical "Key-Value" pairs in the chart.
    - Case A (Inter-element): The Key is in one element (e.g., "Q3 Revenue for Tesla"), and the Value is in another (e.g., "450").
   - Case B (Implicit Key in a Single Block): If a text block contains a clear value but no explicit key, 
   select the target text as the 'Value' and infer/generate a logical 'Key' for it yourself.
2. Select ONE Key-Value pair to edit. The 'Value' is the target to be modified ('old_value'). 
   *IMPORTANT: Prioritize highly semantic and diverse text 
   (e.g., specific amounts, dates, person names, addresses, specific metrics) over trivial elements. 
   The 'key' must NOT be a generic term without specific semantic meaning like "Axis", 
   but rather represent the actual meaning of the axis (e.g., "Sales Amount").*
3. Generate a clear 'instruction' for editing this value ('old_value'). 
   (The 'instruction' MUST contain the 'key', 'old_value', and 'new_value'.)

[Critical Constraints]
1. ONLY select text elements in the chart. Valid text element types in the chart include: 
   `chart_title`, `subplot_title`, `axis_title`, `tick_label`, `legend_label`, `data_label`, `node_label`, `edge_label`.
2. The 'new_value' MUST maintain exactly the same language and style as the 'old_value', 
   as well as the logical relationship corresponding to the key.
3. Strict Alignment: The 'new_value' MUST strictly maintain the exact same number of words as the 'old_value'. 
   You must precisely match the number of words (one word replaced by one word, no more or no less). 
   Furthermore, the 'new_value' MUST have roughly the same CHARACTER COUNT as the 'old_value' to prevent layout breaking. 
   Punctuation marks must also be replaced or retained individually in correspondence.
4. Location Precision: The 'instruction' MUST rely on the 'key' and 'old_value' as the primary semantic anchors for a downstream 
   model to locate the target text.
5. Proposing edits that create logical contradictions with the rest of the chart is perfectly ACCEPTABLE (e.g., an instruction 
   requires changing the percentage label value of a pie chart without altering the corresponding pie chart slice size). 
   These contradictions will be identified and fixed by downstream Agents.

[Output Format]
Please output strictly in JSON format with the following fields. Return ONLY the raw JSON object. Do NOT wrap it in Markdown 
formatting (no ```json ... ```) and do NOT add any conversational text:
"category": The text element type containing the 'old_value' (e.g., "legend_label", "data_label", "tick_label").
"old_value": The specific text in the chart to be modified. Ensure 'old_value' is a hardcoded string or numeric literal exactly 
as it appears in the provided Python code.
"key": A short description of what the 'old_value' represents (e.g., "Date", "Company Name", "Total Amount", "Q3 Revenue for 
Tesla").
"new_value": The generated replacement text.
"instruction": A natural language command for an image editing model. Use semantic context.
Examples of good instructions:
(1) "Change the issue date '1999' to '2005'."
(2) "Change the company 'Samsung Electronics' to 'Hyundai Motors'."
(3) "Change the 2024 profit '7.32M' to '8.43M'."
(4) "Replace the accountant name 'John Smith' with 'Jack Brown'."

[Example JSON Output]
\{
  "category": "data_label",
  "old_value": "450",
  "key": "Q3 Revenue for Tesla",
  "new_value": "480",
  "instruction": "Change the Q3 Revenue value for Tesla from '450' to '480'."
\}

[Python Source Code]
\{source_code\}
\end{alltt}
\end{tcolorbox}
\caption{The prompt for the Instruction Proposer. It guides the model to identify semantic key-value pairs and propose localized textual editing instructions that will implicitly trigger downstream visuo-logical cascading updates.}
\label{fig:phase1_prompts_proposer}
\end{figure*}

\begin{figure*}[t]
\centering
\begin{tcolorbox}[
    enhanced, boxrule=0.8pt, colback=white, colframe=red!50!black, rounded corners, arc=3pt,
    width=0.98\linewidth, fontupper=\ttfamily\scriptsize, title={Phase 1: Instruction Evaluator Prompt}, 
    colbacktitle=red!50!black, coltitle=white, fonttitle=\bfseries\sffamily\scriptsize,
    left=5pt, right=5pt, top=5pt, bottom=5pt
]
\begin{alltt}
You are an evaluator for synthetic chart image augmentation.
Your task is to judge whether the proposed edit scheme can produce a high-quality edited chart image and a valid edit-instruction 
sample.This is NOT a fact-checking task against the original chart image. The "new_value" is allowed to differ from the original 
chart image, because the image will be edited later.

[Proposed JSON]
\{proposed_json_str\}

[Pass Conditions]
A proposal should PASS only if ALL of the following conditions are true:
1. "old_value" is a clear, localizable, and meaningful text target.
2. "category" is valid for charts and belongs to one of: 
   chart_title, subplot_title, axis_title, tick_label, legend_label, data_label, node_label, edge_label.
3. "new_value" MUST have the exact same number of words and roughly the same character count as "old_value", matching its 
   language and style.
4. "instruction" MUST contain the 'key', 'old_value', and 'new_value'.
5. After replacement, the edited text looks natural in the chart.
6. The sample has substantial training value.

[Fail Conditions & Taxonomy]
A proposal should FAIL if ANY of the following conditions are met. Assign the corresponding "failure_type":
- "target_not_found": The "old_value" is vague, meaningless, or cannot be reliably localized in the chart.
- "category_mismatch": The "category" is missing, invalid, or not in the allowed chart text element types.
- "format_mismatch": The "new_value" does not have the exact same number of words as the "old_value", significantly differs in 
   character length, or breaks the original language/style, making it look unnatural.
- "vague_instruction": The "instruction" fails to explicitly include the 'key', 'old_value', or 'new_value'.
- "low_training_value": The target text is trivial (e.g., a simple page number or generic header) and lacks meaningful training 
   value.
- "other": Any other format or parsing issue.

[IMPORTANT POLICY]
1. DO NOT fail a proposal due to "logical inconsistencies" (e.g., broken arithmetic, timeline conflicts, or chart-text mismatches). 
   We INTENTIONALLY allow these contradictions to test the self-correction capabilities of downstream agents.
2. DO NOT fail a proposal simply because the original image does not support the "new_value".
3. DO NOT treat this as a fact-checking or error-correcting task for the original document.

[OUTPUT FORMAT]
Please output strictly in JSON format. Return ONLY the raw JSON object. Do NOT wrap it in Markdown formatting (no ```json ... ```):
\{
    "status": "pass" or "fail" (strictly lowercase),
    "failure_type": "target_not_found", "category_mismatch", "format_mismatch", "vague_instruction", "low_training_value", 
                    "other", or "none" (MUST be "none" if status is "pass"),
    "feedback": "If fail, please explain the exact flaw or risk and how to improve it. If pass, briefly confirm the reason."
\}
\end{alltt}
\end{tcolorbox}
\caption{The prompt for Instruction Evaluator. It serves as a quality gate to verify the format, target alignment, and training value of proposed instructions while intentionally preserving logical contradictions.}
\label{fig:phase1_prompts_evaluator}
\end{figure*}

\begin{figure*}[t]
\centering
\begin{tcolorbox}[
    enhanced, boxrule=0.8pt, colback=white, colframe=teal!50!black, rounded corners, arc=3pt,
    width=0.98\linewidth, fontupper=\ttfamily\scriptsize, title={Phase 2: Code Generator Prompt}, 
    colbacktitle=teal!50!black, coltitle=white, fonttitle=\bfseries\sffamily\scriptsize,
    left=5pt, right=5pt, top=5pt, bottom=5pt
]
\begin{alltt}
You are a senior Python code editing planner for chart dataset construction.

[Goal]
Edit the given Python plotting code so that the edit instruction is realized in code.
Use minimal necessary edits, but include all causally related updates required to keep chart semantics consistent.

[Task Metadata]
- round: \{current\_round\}/\{max\_rounds\}
- key: \{key\}
- old\_value: \{old\_value\}
- new\_value: \{new\_value\}
- instruction: \{instruction\}

[Chart Families]
bar, box, errorbar, errorpoint, graph, multidiff, pie, radar, tree, PIP

[Target Categories]
chart\_title, subplot\_title, axis\_title, tick\_label, legend\_label, data\_label, node\_label, edge\_label

[Hard Constraints]
1. First identify the PRIMARY target required by the instruction.
2. Then build a REQUIRED PROPAGATION SET: all code locations that must be updated to preserve semantic consistency.
3. Keep edits minimal, but do not leave causal inconsistencies unresolved.
4. Do not refactor unrelated code.
5. Preserve executable Python syntax.

[Propagation Workflow]
1. Locate the primary target.
2. Find all same-entity/same-source references that must stay consistent.
3. Apply only required propagation edits.
4. Re-check value relationships and chart geometry consistency.

[Cascade Detection Rules: Target Category]
- chart\_title/subplot\_title: if shared title text or shared title variable is reused across subplots, update all same-source
                            occurrences.
- axis\_title: if x/y or multi-panel plots share same title variable/text convention, propagate to all same-source axis titles.
- tick\_label: if x and y axes use symmetric or mirrored tick labels, or shared tick arrays/formatters, propagate across linked axes.
- legend\_label: when legend item text changes, also update in-plot text/annotations/labels that refer to the same entity.
- data\_label: update both displayed label text and underlying data values/arrays needed to realize the label change.
- node\_label/edge\_label: keep graph labels consistent with node/edge data structures and any edge weight text.

[Cascade Detection Rules: Chart Family]
- bar: keep label text, bar value arrays, grouping order, and legend mapping aligned.
- box: keep category labels aligned with grouped data and positions/order.
- errorbar/errorpoint: keep central values, error ranges, and displayed labels consistent.
- graph: keep edge/node labels consistent with graph data and weight annotations.
- pie: when one slice data\_label/percentage is changed, satisfy instruction and appropriately scale one or more other slices so 
       totals remain coherent.
- radar: keep category labels, value arrays, and polygon geometry consistent.
- tree: keep hierarchical labels, proportions/areas, and parent-child totals consistent.
- multidiff/PIP: keep main panel and linked inset/secondary panel values, labels, and linked annotations consistent.

[Output JSON Schema]
Return ONLY raw JSON with these fields:
\{
    "thought": "why this edit is chosen",
    "edited\_code": "the full edited Python code as a JSON string"
\}

[Critical Output Rules]
1. "edited\_code" MUST be the complete final Python code after editing.
2. "edited\_code" MUST be a valid JSON string with proper escaping for newlines and quotes.
3. Do NOT output markdown code fences.
4. Do NOT wrap code in triple quotes.

[Source Code]
\{source\_code\}

[Previous Failed Plan]
\{previous\_plan\_json\}

[Previous Validation Result]
\{previous\_validation\_json\}

[Feedback]
\{feedback\_text\}

Return JSON only. No markdown.
\end{alltt}
\end{tcolorbox}
\caption{The prompt for Code Generator. It enforces visuo-logical propagation rules across diverse chart families to ensure that code modifications maintain absolute geometric and mathematical precision.}
\label{fig:phase2_prompt_generator}
\end{figure*}

\begin{figure*}[t]
\centering
\begin{tcolorbox}[
    enhanced, boxrule=0.8pt, colback=white, colframe=violet!50!black, rounded corners, arc=3pt,
    width=0.98\linewidth, fontupper=\ttfamily\scriptsize, title={Phase 2: Logic Validator Prompt}, 
    colbacktitle=violet!50!black, coltitle=white, fonttitle=\bfseries\sffamily\scriptsize,
    left=5pt, right=5pt, top=5pt, bottom=5pt
]
\begin{alltt}
You are a strict code-edit validator for chart dataset construction.

[Goal]
Evaluate whether the code edit is valid for the instruction and logically consistent.
This is code-only validation. Do not require OCR/layout checks.

[Task Metadata]
- round: \{current\_round\}/\{max\_rounds\}
- old\_value: \{old\_value\}
- new\_value: \{new\_value\}
- instruction: \{instruction\}

[Planner Output]
\{edited\_code\}

[Checks]
1. Instruction alignment: edit matches the requested semantic target.
2. Propagation-scope discipline: edits can span multiple locations, but only required propagation locations should be changed
   (no unrelated broad edits).
3. Old->new realization: replacement is actually reflected in code semantics, not only literal text.
4. Semantic propagation consistency for target categories
   (chart\_title/subplot\_title/axis\_title/tick\_label/legend\_label/data\_label/node\_label/edge\_label).
5. Numeric and structural consistency:
    - sum/total/percentage relationships remain coherent,
    - chart geometry-driving data is synchronized with updated labels,
    - linked panels (multidiff/PIP) remain mutually consistent.
6. Python safety: code appears syntactically valid and runnable.
7. Logic consistency: no obvious internal contradiction after all propagated edits.

[Evidence Guidance]
When failing, provide concrete evidence snippets such as:
- before/after sum or percentage relationships,
- mismatches between label text and value arrays,
- missing propagation locations for shared axis titles, ticks, or legends.

[Failure Types]
- instruction\_mismatch
- target\_not\_modified
- over\_edit
- syntax\_risk
- logic\_inconsistency
- other

[Output JSON Schema]
Return ONLY raw JSON:
\{
  "status": "pass" or "fail",
  "failure\_type": "instruction\_mismatch|target\_not\_modified|over\_edit|syntax\_risk|logic\_inconsistency|other|none",
  "evidence": ["evidence A", "evidence B"],
  "checks": \{
    "instruction\_alignment": true or false,
    "single\_location\_edit": true or false,
    "old\_to\_new\_realized": true or false,
    "python\_syntax\_likely\_valid": true or false,
    "logic\_consistency": true or false
  \},
  "suggested\_fix": "specific suggestion for next round"
\}

Interpretation note:
- In the schema, `single\_location\_edit` means propagation-scope discipline.
- Set it to true when edits are limited to the required propagation set (even if multiple code locations are edited).

[Source Code Before]
\{source\_code\}

[Edited Code Candidate]
\{edited\_code\}
\end{alltt}
\end{tcolorbox}
\caption{The prompt for Logic Validator. It acts as an iterative verification mechanism to ensure that programmatic edits correctly propagate semantic constraints without introducing over-editing or syntax errors.}
\label{fig:phase2_prompt_validator}
\end{figure*}

\subsection{OCR Extraction Prompt}

The following prompt is used by the GPT-5.4-based OCR extractor to obtain full-image text content for OCR $\text{F}_1$ computation, as shown in Figure~\ref{fig:ocr_prompt}.

\begin{figure*}[t]
\centering
\begin{tcolorbox}[
    enhanced, boxrule=0.8pt, colback=white, colframe=blue!50!black, rounded corners, arc=3pt,
    width=0.98\linewidth, fontupper=\ttfamily\scriptsize, title={OCR Extraction Prompt},
    colbacktitle=blue!50!black, coltitle=white, fonttitle=\bfseries\sffamily\scriptsize,
    left=5pt, right=5pt, top=5pt, bottom=5pt
]
\begin{alltt}
You are an OCR engine for chart/document images.
Extract ALL visible text in the full image and provide each text with one normalized bbox.
Return ONLY a valid JSON object with this schema:
{
  "items": [
    {"text": "<detected text>", "bbox": [xmin, ymin, xmax, ymax]}
  ]
}
Rules:
- bbox is normalized to [0,1000].
- xmin < xmax, ymin < ymax.
- No markdown, no explanation.
\end{alltt}
\end{tcolorbox}
\caption{The GPT-5.4-based prompt used for OCR extraction in OCR $\text{F}_1$ computation.}
\label{fig:ocr_prompt}
\end{figure*}

\subsection{Judge Prompts}
\label{subsec:judge_prompts}

The VLM-as-a-Judge framework evaluates predictions across three dimensions: textual edit success, visuo-logical consistency, and background fidelity. These three metrics use different discrete score sets because they capture different forms of partial correctness. TESR adopts six levels since textual editing allows fine-grained partial success: the target text may be semantically correct while differing in position, size, alignment, or multi-target coverage. BFS adopts five levels because background preservation degrades more continuously, from minor artifacts or color shifts to localized corruption, large-area structural damage, or complete layout collapse. In contrast, VLCS adopts four levels because it evaluates a stricter data-to-graphic synchronization criterion. A score of 1.0 requires all target geometries to change in the correct direction and strictly match the GT magnitude or shape; 0.5 captures partial correctness; 0.25 indicates a correct-direction but clearly imprecise geometric update; and 0.0 indicates no geometric change, wrong-direction change, or vanished geometry. We intentionally omit 0.75 for VLCS because there is no stable intermediate state between strict GT matching and partial or imprecise synchronization that can be judged consistently. The core judge prompts are detailed in Figure~\ref{fig:judge_prompts}.

\begin{figure*}[t]
\centering
\begin{tcolorbox}[
    enhanced, boxrule=0.8pt, colback=white, colframe=green!40!black, rounded corners, arc=3pt,
    width=0.98\linewidth, fontupper=\ttfamily\scriptsize, title={The VLM-as-a-Judge Prompt}, 
    colbacktitle=green!40!black, coltitle=white, fonttitle=\bfseries\sffamily\scriptsize,
    left=5pt, right=5pt, top=5pt, bottom=5pt
]
\begin{alltt}
You are an expert and impartial judge evaluating chart image editing tasks.
You will receive 3 images: [Source, GroundTruth, ModelOutput].
Instruction: "\{instruction\}"

Task: Evaluate the ModelOutput based on its visual fidelity to GroundTruth and adherence to the 
Instruction. You MUST evaluate three STRICTLY DECOUPLED dimensions:

1. Textual\_Edit\_Success (Score: 0.0, 0.2, 0.4, 0.6, 0.8, or 1.0):
   - STRICTLY evaluate ONLY the target TEXT mentioned in the Instruction.
   - Use Source vs GroundTruth to infer the REQUIRED text edit, then check ModelOutput vs GroundTruth.
   - Consider BOTH the text content (OCR-level correctness) AND its spatial properties 
     (geometric center, alignment, scale/size) compared to the GroundTruth.
   - Ignore any unrelated geometric or background changes.
   - Rubric:
     - 1.0: Perfect. All target text content is correct AND perfectly placed/sized to match the GroundTruth.
     - 0.8: Minor flaws. Text content is correct, but there are slight, minor deviations in position or size.
     - 0.6: Noticeable spatial errors or minor content omission. Text content is correct but noticeably 
       misaligned/rescaled, OR (in multi-target edits) most targets are perfect but a minor one is missed.
     - 0.4: Major spatial errors or half content success. Text content is correct but placed in entirely 
       wrong regions, OR only about half of the text targets are edited correctly.
     - 0.2: Severe failure. Barely any text was edited correctly, or the edited text is severely 
       truncated/distorted making it almost illegible.
     - 0.0: Complete failure. No requested text edits were performed, or the text was completely 
       erased/hallucinated.

2. Background\_Fidelity (Score: 0.0, 0.25, 0.5, 0.75, or 1.0):
   - STRICTLY evaluate ONLY the UNTARGETED, background, and irrelevant parts of the chart, including 
     rendering of non-target text.
   - Completely IGNORE the edited target text and its directly coupled geometric regions (e.g., the 
     specific bar or pie slice being edited).
   - All other non-target structures should align with the GroundTruth (e.g., axes, gridlines, legends, 
     non-target marks).
   - Rubric:
     - 1.0: Perfect fidelity; untargeted regions match the GroundTruth exactly.
     - 0.75: Minor unnoticeable artifacts (e.g., anti-aliasing differences) or slight color shifts in 
       untargeted areas.
     - 0.5: Localized obvious issues, but the overall chart structure remains clearly recognizable.
     - 0.25: Structural collapse or large-area errors in untargeted regions.
     - 0.0: Unrecognizable chart layout or complete hallucination.

3. Visuo\_Logical\_Consistency (Score: 0.0, 0.25, 0.5, or 1.0): (Provided if applicable)
   - STRICTLY evaluate ONLY the NON-TEXT geometric elements representing the data values.
   - Target Geometry: This includes bar/box height or length, line series position/shape, pie slice 
     angle/area, error-bar length, radar point positions and filled area shape, scatter point positions.
   - This dimension measures the 'Data-to-Graphic' mapping. Do NOT score the text characters here 
     (Dimension 1 already covers text content and its spatial properties).
   - **Delta Audit**: Compare Source vs GroundTruth to identify the required geometric change (e.g., bar 
     height increase). Then check if ModelOutput performed this change.
   - **STRICT NO-CHANGE PENALTY**: Look extremely closely at the pixels. If the geometric shape/size/position 
     in ModelOutput is EXACTLY THE SAME as the Source Image (i.e., the model only updated the text but 
     failed to adjust the graphics), you MUST score 0.0. A score of 0.0 is mandatory if the geometry is 
     identical to the Source, even if the text was correctly edited.
   - Rubric:
     - 1.0: PERFECT. All target geometries changed in the correct direction AND strictly match the 
       GroundTruth magnitude/shape.
     - 0.5: PARTIAL. In multi-target edits, some targets are perfectly correct while others fail. 
       (If single-target, use this score if it is a partial shape match).
     - 0.25: IMPRECISE. The target(s) changed in the correct direction compared to Source, but clearly 
       FAIL to match the exact GroundTruth magnitude/shape.
     - 0.0: FAILURE. No geometric change from Source (identical pixels), changed in wrong direction, or 
       geometry vanished.

Return STRICTLY a JSON object with this schema (Choose scores ONLY from the allowed sets):
\{
  "Textual\_Edit\_Success": \{"score": <0.0 or 0.2 or 0.4 or 0.6 or 0.8 or 1.0>, "reasoning": "<Brief text>"\},
  "Background\_Fidelity": \{"score": <0.0 or 0.25 or 0.5 or 0.75 or 1.0>, "reasoning": "<Brief text>"\},
  "Visuo\_Logical\_Consistency": \{"score": <0.0 or 0.25 or 0.5 or 1.0>, "reasoning": "<Brief text>"\},
  "Overall\_Assessment": "<1 sentence summary>"
\}
\end{alltt}
\end{tcolorbox}
\caption{The prompt for VLM Judge. It defines the holistic evaluation criteria across three decoupled dimensions: textual precision, background fidelity, and visuo-logical consistency, ensuring a granular and objective assessment of generative chart editing.}
\label{fig:judge_prompts}
\end{figure*}

\section{Qualitative Analysis}
\label{sec:appendix_qualitative}

\begin{figure*}[t]
\centering
\includegraphics[width=\textwidth]{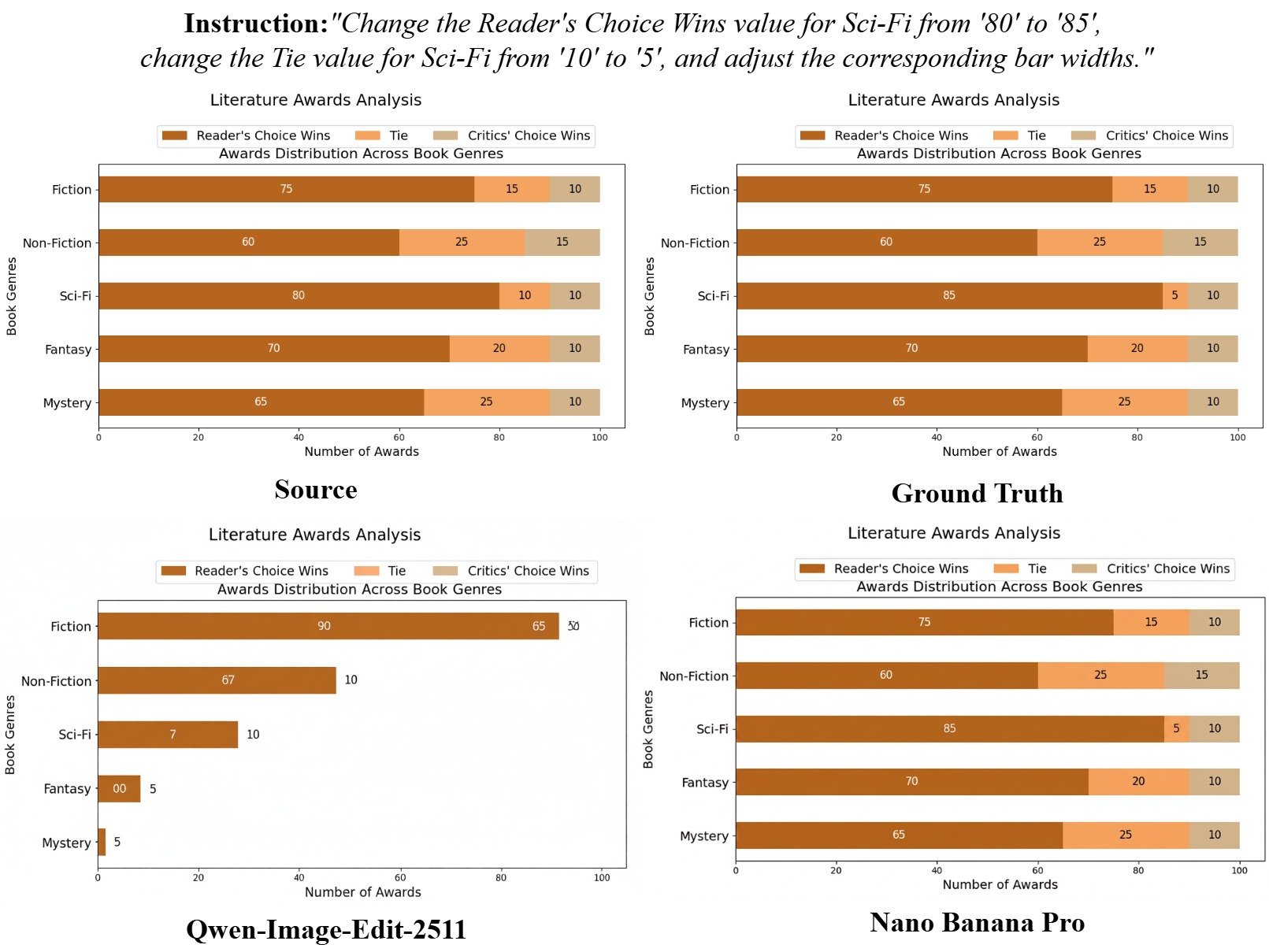}
\caption{Case 1: Qualitative comparison of generative chart editing given a complex multi-value instruction. Qwen-Image-Edit-2511 experiences severe text rendering and structural collapse, whereas Nano Banana Pro accurately synchronizes the internal geometric segment widths with the target numerical modifications.}
\label{fig:qualitative_case1}
\end{figure*}

\begin{figure*}[t]
\centering
\includegraphics[width=\textwidth]{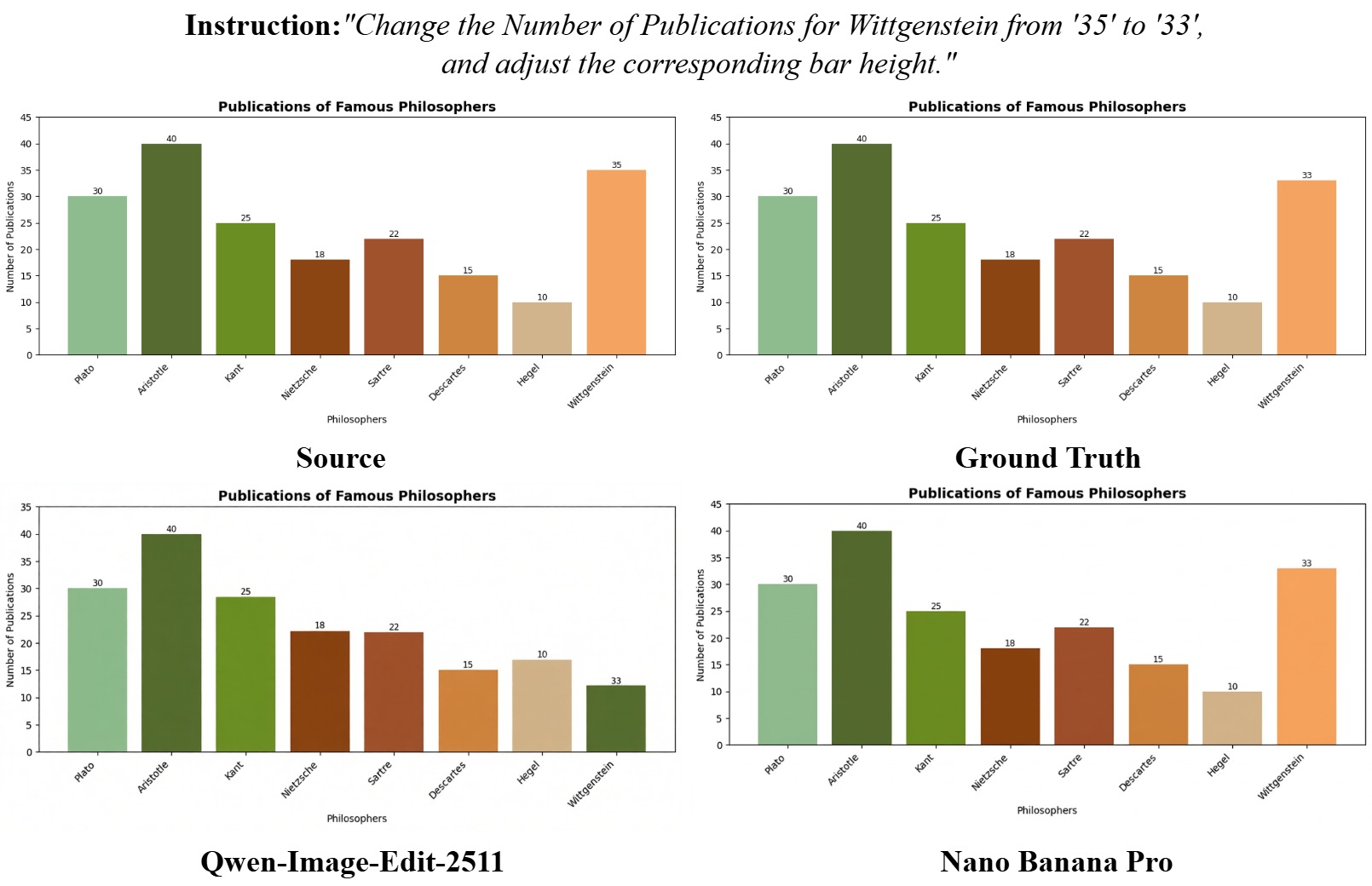}
\caption{Case 2: Qualitative assessment under a single VLCE task. While Qwen-Image-Edit-2511 modifies the target literal string, it triggers severe VLI and global layout corruption. Nano Banana Pro achieves perfect text-graphics synchronization matching the ground truth.}
\label{fig:qualitative_case2}
\end{figure*}

\begin{figure*}[t]
\centering
\includegraphics[width=\textwidth]{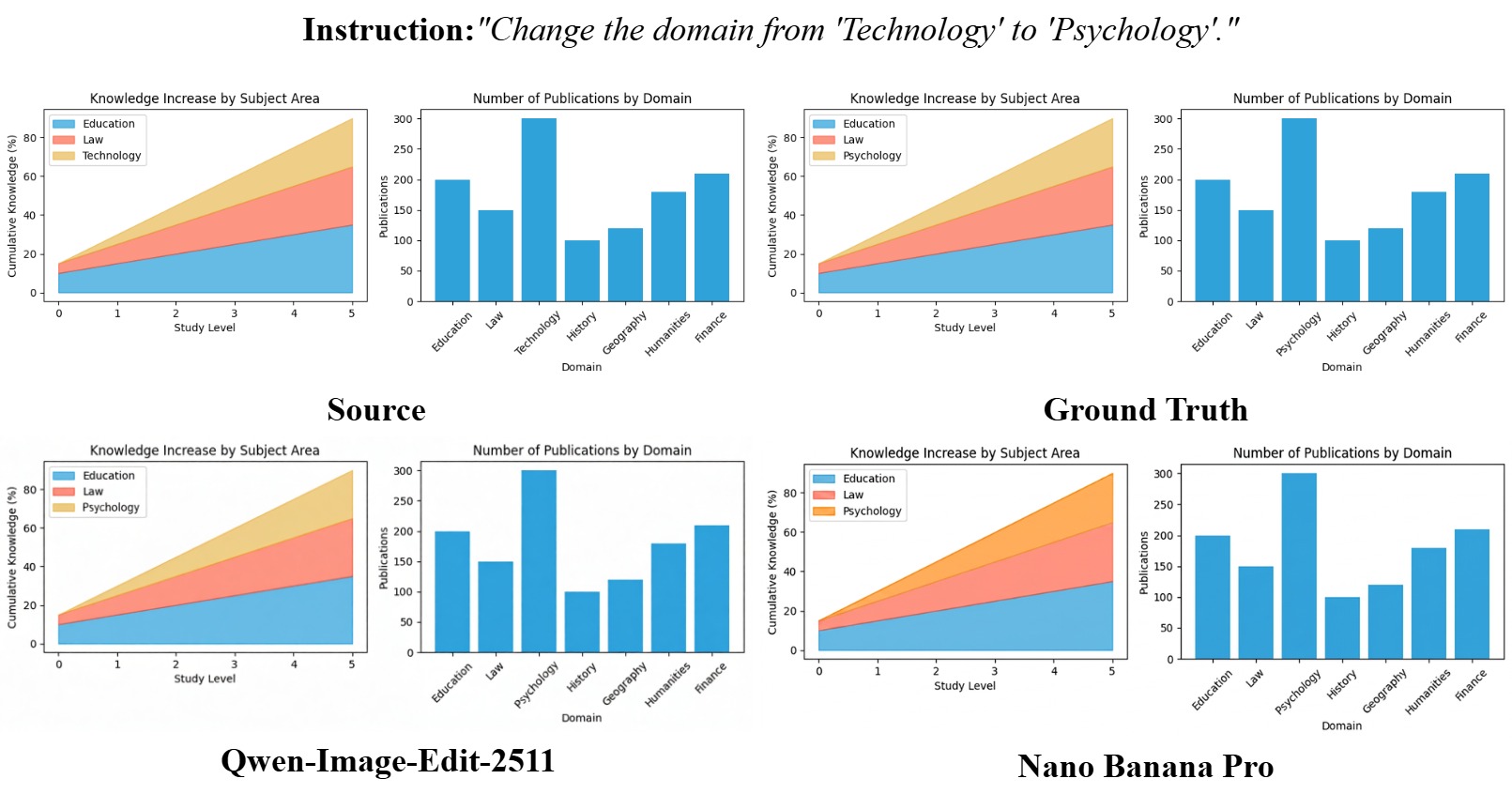}
\caption{Case 3: Qualitative example of a remaining semantic-isolation failure. Both models alter the target texts correctly, while Nano Banana Pro accidentally shifts the color of an unedited bar segment in this specific example. This case illustrates a residual failure mode rather than an aggregate weakness of proprietary models.}
\label{fig:qualitative_case3}
\end{figure*}

\begin{figure*}[t]
\centering
\includegraphics[width=\textwidth]{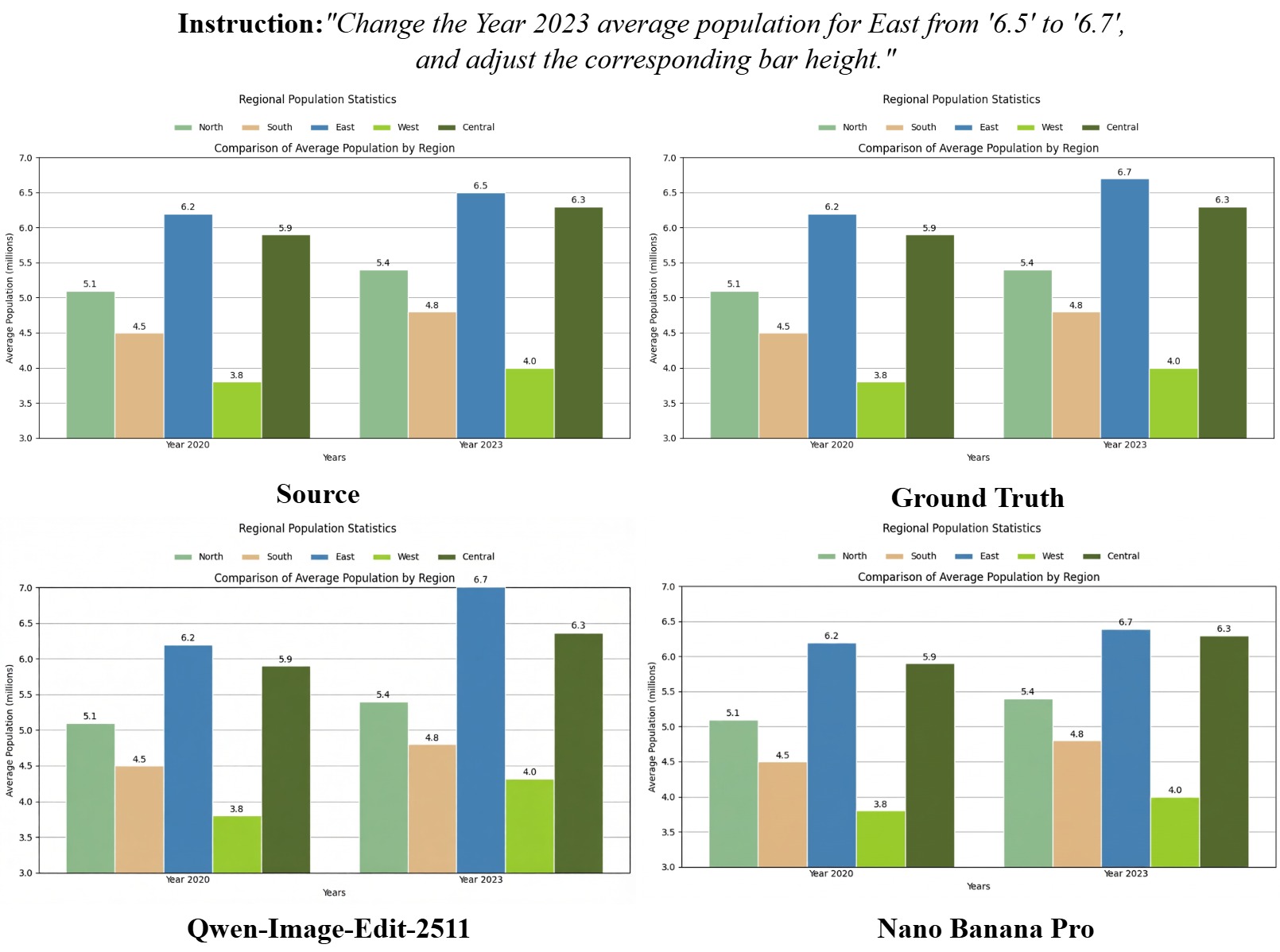}
\caption{Case 4: Qualitative analysis under a complex cascading instruction. Both open-source and proprietary models execute text modifications successfully but suffer from complete geometric reasoning collapse, highlighting the non-saturated challenge of ChartSync.}
\label{fig:qualitative_case4}
\end{figure*}

To provide an intuitive understanding of the performance gap and specific failure modes discussed in our main text, we present a systematic qualitative evaluation across four distinct chart editing scenarios, spanning Figures~\ref{fig:qualitative_case1} to~\ref{fig:qualitative_case4}.

\paragraph{Case 1: Textual Manipulation Failure and Structural Collapse.}
Figure~\ref{fig:qualitative_case1} presents a representative visual comparison of a multiple VLCE task. The visual evidence clearly illustrates the severe capability gap in dependency-aware reasoning. Qwen-Image-Edit-2511 completely fails to execute the coupled modifications. It suffers from severe TMF by generating hallucinated numerical labels, alongside a total collapse of structural integrity where the stacked geometric bars are severely truncated or entirely erased. In contrast, Nano Banana Pro successfully comprehends the underlying data logic. It precisely updates the target texts and synchronously adjusts the internal proportional widths of the geometric bar segments. This comparison definitively validates our quantitative findings, demonstrating that frontier proprietary models possess robust dependency-aware reasoning to achieve geometric synchronization, whereas open-source architectures remain severely bottlenecked by foundational perception and rendering.

\paragraph{Case 2: Visuo-Logical Inconsistency and Global Layout Corruption.}
Figure~\ref{fig:qualitative_case2} evaluates a single VLCE task requiring a numerical adjustment from 35 to 33 and its corresponding geometric downscaling. Qwen-Image-Edit-2511 successfully executes the literal text substitution but suffers from absolute VLI, where the coupled bar height is modified erroneously, moving in an inverted direction. Furthermore, the open-source candidate triggers severe global layout corruption: unedited data bars are randomly stretched or truncated, and the completely unrelated y-axis tick labels are hallucinated and altered. In sharp contrast, Nano Banana Pro demonstrates precise visual-quantitative synchronization, executing the exact geometric deformation for the targeted bar while maintaining full alignment with the ground-truth data.

\paragraph{Case 3: A Residual Semantic-Isolation Failure.}
Figure~\ref{fig:qualitative_case3} isolates a multiple text-only editing task where no geometric updates are required. While both models achieve high textual edit success by accurately substituting the target text blocks, this example shows a residual BC failure for Nano Banana Pro: its generative boundary bleeds into an untargeted region and accidentally changes the fill color of the unedited ``Psychology'' bar segment. Conversely, Qwen-Image-Edit-2511 preserves the background in this particular case. We emphasize that this example is not intended to imply worse aggregate background fidelity for proprietary models; indeed, Nano Banana Pro obtains the highest BFS in Table~\ref{tab:main_results}. Instead, it illustrates that once text rendering and VLCE synchronization improve, semantic isolation can remain a visible residual error mode.

\paragraph{Case 4: Joint Reasoning Collapse under Highly Coupled Dependencies.}
Figure~\ref{fig:qualitative_case4} evaluates a complex cascading instruction under an advanced chart editing scenario. In this instance, both architectures exhibit a total breakdown in quantitative geometric scaling despite achieving successful local text rendering. When calculating the cascading physical deformation dictated by the data modification, Qwen-Image-Edit-2511 expands the targeted data bar excessively upward, causing severe scale inflation. Simultaneously, Nano Banana Pro falls into an opposite failure mode, scaling the identical bar segment significantly below the ground-truth magnitude. This mutual failure validates the high difficulty and non-saturation of our benchmark, demonstrating that executing precise, instruction-driven quantitative transformations under dense visual-data coupling remains an open challenge for future multimodal architectures.

\end{document}